\documentclass{Interspeech2024}

\interspeechcameraready

\title{Target conversation extraction: Source separation using turn-taking dynamics}

\name{Tuochao Chen,$^1$ Qirui Wang,$^1$ Bohan Wu,$^1$ Malek Itani,$^1$ Sefik Emre Eskimez,$^2$ Takuya Yoshioka,$^3$  Shyamnath Gollakota$^1$ }

\address{$^1$University of Washington, $^2$Microsoft, $^3$AssemblyAI}
\email{\{tuochao,qw43,bohanw,malek\}@cs.washington.edu, sefik.eskimez@microsoft.com, takuya.yoshioka@ieee.org, gshyam@cs.washington.edu}


\begin{document}

\maketitle

\begin{abstract}
Extracting the speech of participants in a conversation amidst interfering speakers and noise presents a challenging problem. In this paper, we introduce the novel task of target conversation extraction, where the goal is to extract the audio of a target conversation based on the speaker embedding of one of its participants. To accomplish this, we propose leveraging temporal patterns inherent in human conversations, particularly turn-taking dynamics, which uniquely characterize speakers engaged in conversation and distinguish them from interfering speakers and noise. Using neural networks, we show  the feasibility of our approach on English and Mandarin conversation datasets. In the presence of interfering speakers, our results show an 8.19~dB improvement in signal-to-noise ratio for 2-speaker conversations and a 7.92~dB improvement for 2-4-speaker conversations.
\noindent {Code, dataset available at {\textcolor{blue}{{{\url{https://github.com/chentuochao/Target-Conversation-Extraction}}}}}}
\end{abstract}

\section{Introduction}



Verbal conversations have long served as a primary mode of human interaction, essential for both sharing information and fostering social connections~\cite{frontiers,turntake1}. While humans can navigate conversations in noisy environments, isolating the audio of a target conversation in the presence of interfering speakers and noise remains a challenging task for machines. 

\begin{figure}[t!]
 \centering
 \includegraphics[width=0.9\linewidth]{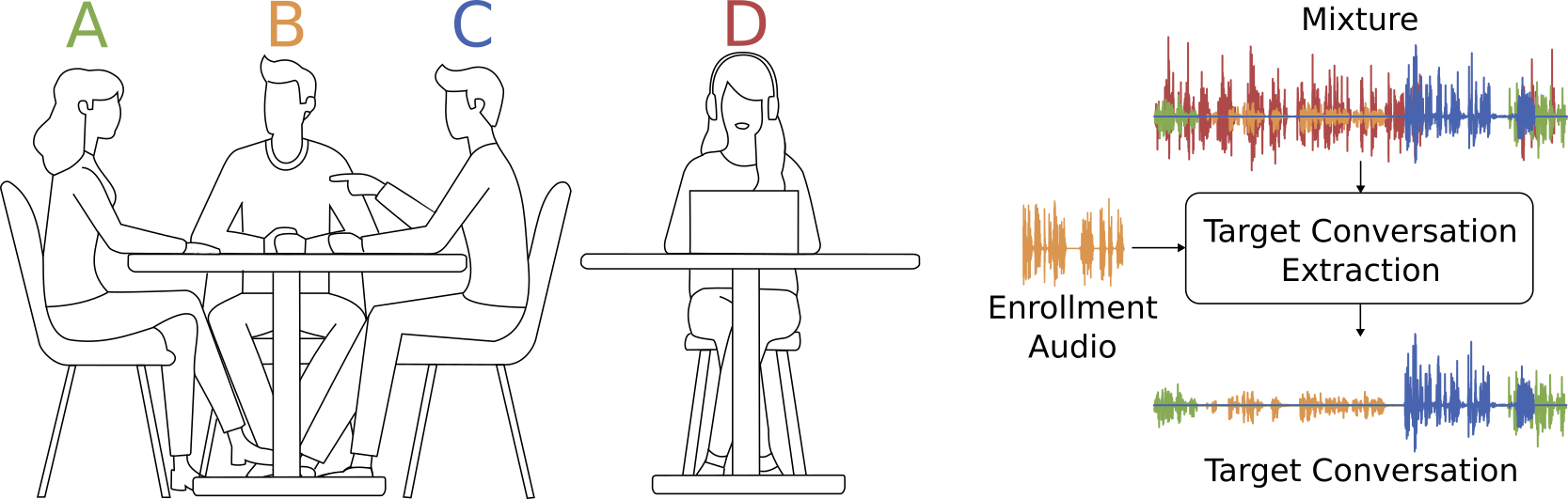}
 \vskip -0.1in
 \caption{The goal of target conversation extraction in this illustration is as follows: given a clean enrollment audio or embedding for B, we want to extract   audio for the conversation between A, B and C, amidst interference from speaker D.}
 \vskip -0.2in
 \label{fig:fig1}
\end{figure}

Consider a scenario in Fig.~\ref{fig:fig1}, where an individual is engaged in a conversation, such as an interview, in a busy cafe. Ideally, the individual's mobile device, equipped with the ability to recognize the owner's voice, should be able to isolate and capture only the voices involved in the conversation, effectively filtering out interfering conversations and noise from nearby sources. This ability to extract target conversations is crucial for both traditional applications like recording interviews in public spaces, enhancing conversations in  multimedia applications, and life logging, as well as for futuristic applications like personalized AI agents that understand, assist, and augment human conversations in the wild.

While existing source separation techniques can extract a target speaker from audio mixtures~\cite{zmolikova}, none have addressed the task of extracting speech from speakers engaged in conversation with a reference speaker. In this paper, we introduce the {\it target conversation extraction (TCE)} task. More formally, our task is to extract the target conversation signal from an audio mixture. This mixture includes the target conversation involving two or more participants, interfering speakers outside the target conversation, and background noise. Our goal is to achieve this extraction based on an enrollment audio or a speaker embedding for one of its participants (e.g., the device owner). 



Our approach for addressing TCE hinges on the crucial role of turn-taking in human conversations~\cite{turntake1}.  Turn-taking dynamics help effectively manage social interactions~\cite{turntake2, turntake3}.  Research  shows  human conversations predominantly feature one speaker at a time, with brief overlaps being common~\cite{frontiers}. However, interfering speakers outside the target conversation do not fit the temporal patterns of the target conversation.



 Therefore, given an enrollment audio or the  corresponding  embedding for a reference speaker (e.g., the device owner), we train a source separation  network to extract speakers adhering to turn-taking dynamics with the reference speaker, while filtering the interfering speakers and background noise. 
 A key challenge for training is the scarcity of open-source, high-quality audio conversational datasets featuring diverse speakers. To tackle this, we propose a timing-preserving data augmentation method for TCE that utilizes large, clean non-conversational speech datasets to augment smaller conversational datasets while preserving the timing  in conversations. 
 
 Evaluations with   RAMC~\cite{yang2022open}, a  2-speaker Mandarin conversation dataset demonstrates a 7.18~dB improvement {in scale-invariant signal-to-distortion ratio (SI-SDR) and 8.19~dB in signal-to-noise ratio (SNR)}, in the presence of two interfering speakers. Additionally, with AMI~\cite{kraaij2005ami}, a 2-4 speaker English conversation dataset with {18.6\%} average overlap between speakers in a conversation, we achieve a 6.32~dB SI-SDR and 7.92~dB SNR improvement with 2-4 interfering speakers. Our augmentation and training approach also enhances separation performance by 2-3 dB. 
 Finally, the model fails when we  alter the speaker timing in the test set to break human turn-taking dynamics,    indicating that our trained separation model can capture these timing patterns.

\section{Related work}

To our knowledge, our work is the first to introduce the TCE task. Here, we describe other tasks related to TCE. 

\noindent\textbf{Target speech extraction:} The goal of this task is to extract a target speaker from a mixture, given auxiliary cues to identify the target~\cite{zmolikova}. These clues can be  audio examples of the target speaker~\cite{i_vector,d_vector,speakerbeam,waveformer,10.1145/3613904.3642057}  or spatial clues~\cite{hybridbeam, gu2019neural,acousticswarm}. Visual clues~\cite{avsepformer},  text queries~\cite{liu2023separate}, and concept embeddings~\cite{conceptbeam} have also been proposed. Our aim instead is to extract a target conversation with multiple speakers, given an enrollment audio for a reference speaker included in the conversation. 

\noindent\textbf{Turn-taking in conversations:} Conversational turn-taking has  temporal patterns that are crucial for understanding human social interaction~\cite{levinson2015timing}. Psycholinguistics literature~\cite{turntake1,turntake2, turntake3} has explored statistical models for conversational turn-taking through corpus analyses. These studies have shown that while conversations mostly have a single speaker talking at a time along with silence and gaps, turn-taking also involves overlaps and backchannels~\cite{turntake1}. Recent works have used deep learning to model and predict~\cite{turn_model,turn_eval,turn_predict} such turn-taking events. 


\noindent\textbf{Conversation-related speech processing:} Speech diarization~\cite{diarization4conversation} is a conversation-related speech task that identifies ``who spoke when". \cite{asr4conversation, diarization4conversation} proposed conversational speech recognition. \cite{qa4conversation} introduced a speech-based question-answering system, while \cite{sa4conversation, sdu4conversation} study conversational status tracking and sentiment analysis. These systems assume that the conversation audio is clean, without interfering speakers.

\section{Methods}

{\bf Problem formulation.} Let $x$ be the noisy conversation signal recording which can be decomposed into three parts:

\begin{align}
x = s_0 + \sum_{i=1}^N s_{i}^{conv} + \sum_{j=1}^M s_{j}^{inter} + n
\end{align}

Here, $s_0$ corresponds to the speech of the reference speaker in the conversation. $s_{1}^{\text{conv}},\ldots,s_{N}^{\text{conv}}$ are the other speakers  in the target conversation,  $s_{1}^{\text{inter}},\ldots,s_{M}^{\text{inter}}$ are the interfering speakers not part of the conversation, and $n$ is background noise.

Given a clean enrollment audio for the reference speaker, we can compute the corresponding speaker embedding of length $K$,  ${\epsilon}_0 \in \mathbb{R}^{K}$, using  a pre-trained speech neural network on the enrollment audio~\cite{d_vector}. The goal of the TCE network $\mathcal{G}$ with parameters $\theta$, then is to extract the  target conversation:
\begin{align}
y = \mathcal{G}(x | \epsilon_0;\theta) =   \hat{s}_0 + \sum_{i=1}^{N} \hat{s}_i^{conv}
\end{align}



\begin{figure}[t!]
  \centering
  \includegraphics[width=0.93\linewidth]{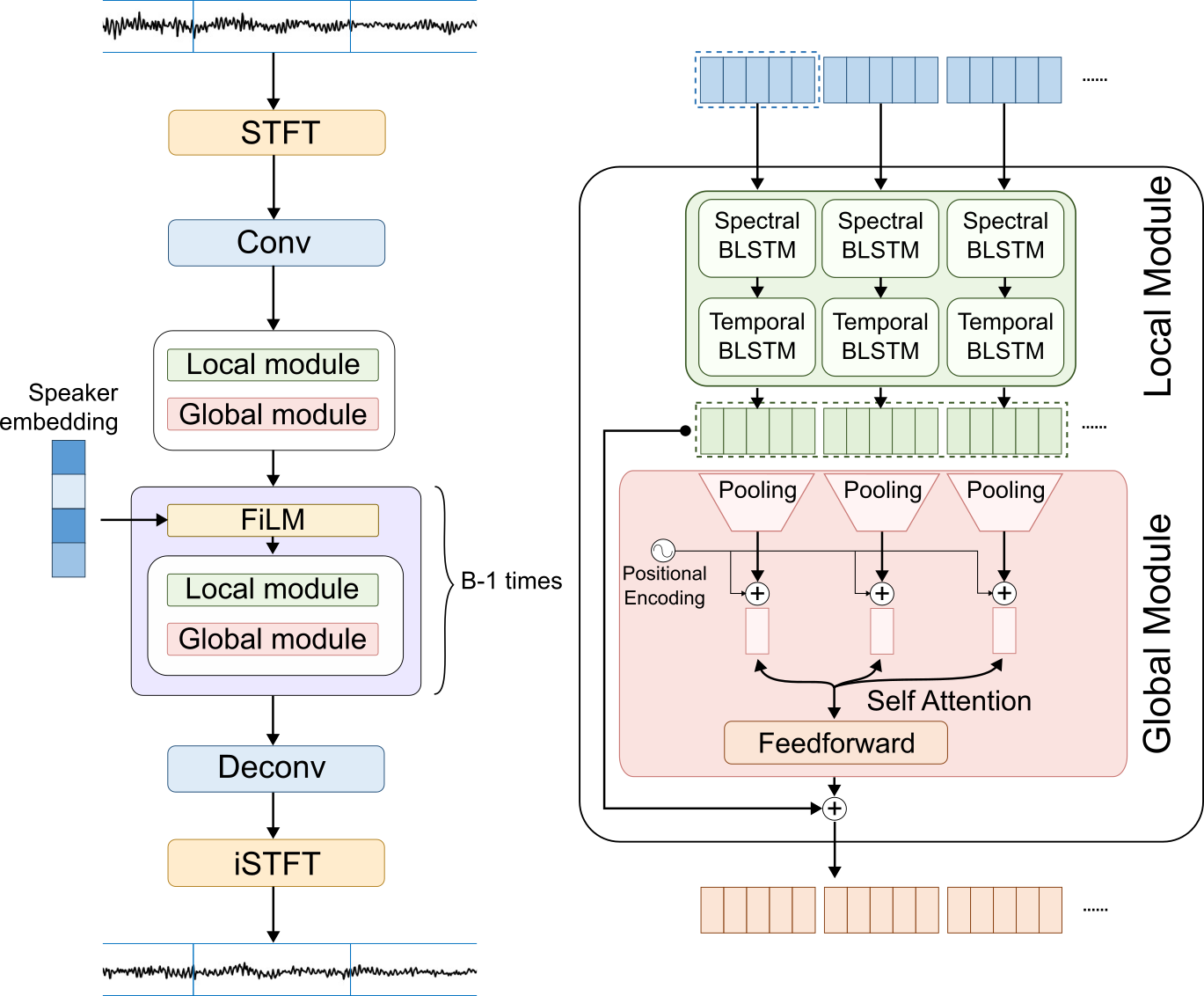}
  \vskip -0.1in
  \caption{Neural network architecture for the TCE task.}
  \vskip -0.2in
  \label{fig:architecture}
\end{figure}

\noindent\textbf{Neural network.} 
We use TF-GridNet~\cite{wang2023tf} as our base   source separation network. It uses dual-path LSTMs across frequency  and time dimensions as well as full attention  on short (1-5s) audio mixtures, for effective source separation~\cite{wang2023tf}. However  traditional recurrent networks face  challenges with long sequences due to parallelization issues~\cite{vaswani2017attention} and are not efficient  for longer minute-long sequences, during both training and inference.  {Further,  full attention time and memory complexity grow quadratically with input length}. So,  our   network, shown in Fig.~\ref{fig:architecture}, 
 employs LSTMs on local audio  chunks followed by sparse pooling attention,  capable of efficiently operating across long sequences~\cite{longformer,poolingformer,resepformer,speechLongformer}. This  enables   source separation while efficiently incorporating turn-taking cues.

Specifically,  we first apply a short-time Fourier transform (STFT)  to obtain the time-frequency (TF) representation $X\in \mathbb{R}^{T\times F}$, where $T$ represents the number of STFT frames and $F$ is the number of frequency bins.  Subsequently, we concatenate the real and imaginary parts of $X$ and feed them into a 2D convolution with a 3 × 3 kernel, resulting in $Y\in \mathbb{R}^{D\times T \times F}$, where $D$ denotes the embedding dimension for each TF bin.  Next, $Y$ undergoes processing through multiple extraction blocks that include FiLM layers~\cite{perez2018film} and local and global modules. The processed TF representation then proceeds through a 2D deconvolution (3 × 3 kernel), followed by an inverse STFT (iSTFT) to recover the time-domain conversation signal.


Each extraction block consists of three modules: a FiLM layer, a local module, and a global module. The FiLM Layer merges the reference speaker embedding, $\epsilon_0$, to the hidden representation in every block except the first one, conditioning the network with the reference speaker. {The local module applies spectral and temporal bidirectional long short-term memory (BLSTM) on local audio chunks. } We employ a sliding window with  size $W$ and stride  $S$ on $Y$ to yield $Z_l \in \mathbb{R}^{C \times D \times W \times F}$ where $C$ represents the number of sliding windows.  Within the local module, like TF-GridNet~\cite{wang2023tf}, we start with a one-layer BLSTM with hidden size $H$, operating on the frequency domain $F$, followed by a deconvolution layer with $2H$ input channels and $D$ output channels. Subsequently, another BLSTM with hidden size $H$ is applied to the time domain, with the sequence length constrained to $W,$ followed by a deconvolution layer.

The global module applies sparse pooling attention across the chunks. We employ a pooling operation with a window size $W$ and stride size $S$ to compress the long sequence $T$ to $C$, yielding $Z_g \in \mathbb{R}^{D \times C \times F}$. Following pooling, we apply positional encoding to $Z_g$ to preserve the order information crucial for conversational turn-taking. Subsequently, three linear layers are utilized to merge the $D$ and $F$ dimensions, resulting in the key $\mathbb{K}\in \mathbb{R}^{C \times E}$, query $\mathbb{Q} \in \mathbb{R}^{C \times E}$, and value $\mathbb{V} \in \mathbb{R}^{C \times (DF/L)}$. Here, $E$ represents the embedding dimension for the key and query, and $L$ denotes the number of heads. Finally, self-attention is applied across the $C$ dimension, followed by a feedforward layer with input and output channel dimensions both set to $D$. The output is then added to the input tensor via a residual connection.

\noindent{\bf Timing-preserving data augmentation and training.}  For training, we require conversation datasets with high-quality audio and clean speech, along with speaker ID and timestamp labels to extract speaker embeddings and avoid speaker repetition.  There are only  a handful of open-source conversation datasets that meet  all these criteria. Further, in some of these datasets, we found that speakers within a conversation often share acoustic properties like reverberations, background noise, and hardware distortion. This can lead the model to rely on these properties rather than turn-taking, when trained just  on this data. 

 Data augmentation methods have been used to improve performance for other speech tasks~\cite{landini2022simulated, yang2023simulating,kanda2022streaming}. For our task,  we leverage non-conversation speech datasets with high-quality speech and diverse speakers to augment conversation data.   By replacing individual speaker segments with speech from these datasets while preserving timing information, we disentangle acoustic properties and encourage the model to learn timing cues. With a probability $p$, each speaker segment in the conversation dataset is replaced with an utterance of the same length from a randomly selected speaker from the non-conversation speech datasets. When $p = 1$, all speakers are replaced, which we term synthetic conversations ($\mathcal{S}$). For $p \in (0, 1)$, only some speakers are replaced, which we term augmented conversations ($\mathcal{A}$). When $p$ is  0, it remains the original real conversations ($\mathcal{R}$). 


  While this augmentation maintains timing information, it strips away prosodic features like pitch, voice quality, and intensity, resulting in unnatural conversations. So, we pre-train the model using synthetic and augmented conversations. By randomly substituting speech segments, the model is encouraged to learn conversational timing patterns. We then fine-tune the model using augmented  and real conversation data. 


 \begin{table}[t!]
  \caption{Results on mixtures of real Mandarin conversations.}
  \vskip -0.1in
  \label{tab:mandarain}
  \centering
  \begin{tabular}{ c c c  }
    \toprule
    \multicolumn{1}{c}{\textbf{Pre-training/fine-tuning}} & 
    \multicolumn{1}{c}{\textbf{SI-SDRi}} &
    \multicolumn{1}{c}{\textbf{SNRi}} \\
    
    \midrule
     $\mathcal{R}_{zh}$  & 4.42  & 6.30 \\
     $\mathcal{R}_{zh}$, $\mathcal{A}_{zh}$  & 5.81  & 7.02\\
     $\mathcal{S}_{cross}, \mathcal{S}_{stats}$/$\mathcal{R}_{zh}$ & 6.96   & 8.02\\
   $\mathcal{S}_{cross}, \mathcal{S}_{stats}$/$\mathcal{R}_{zh}$, $\mathcal{A}_{zh}$   & \textbf{7.18}   &  \textbf{8.19} \\
    \bottomrule
  \end{tabular}
    \vskip -0.1in
\end{table}

\begin{figure}[t]
  \centering
  \vskip -0.1in
  \includegraphics[width=\linewidth]{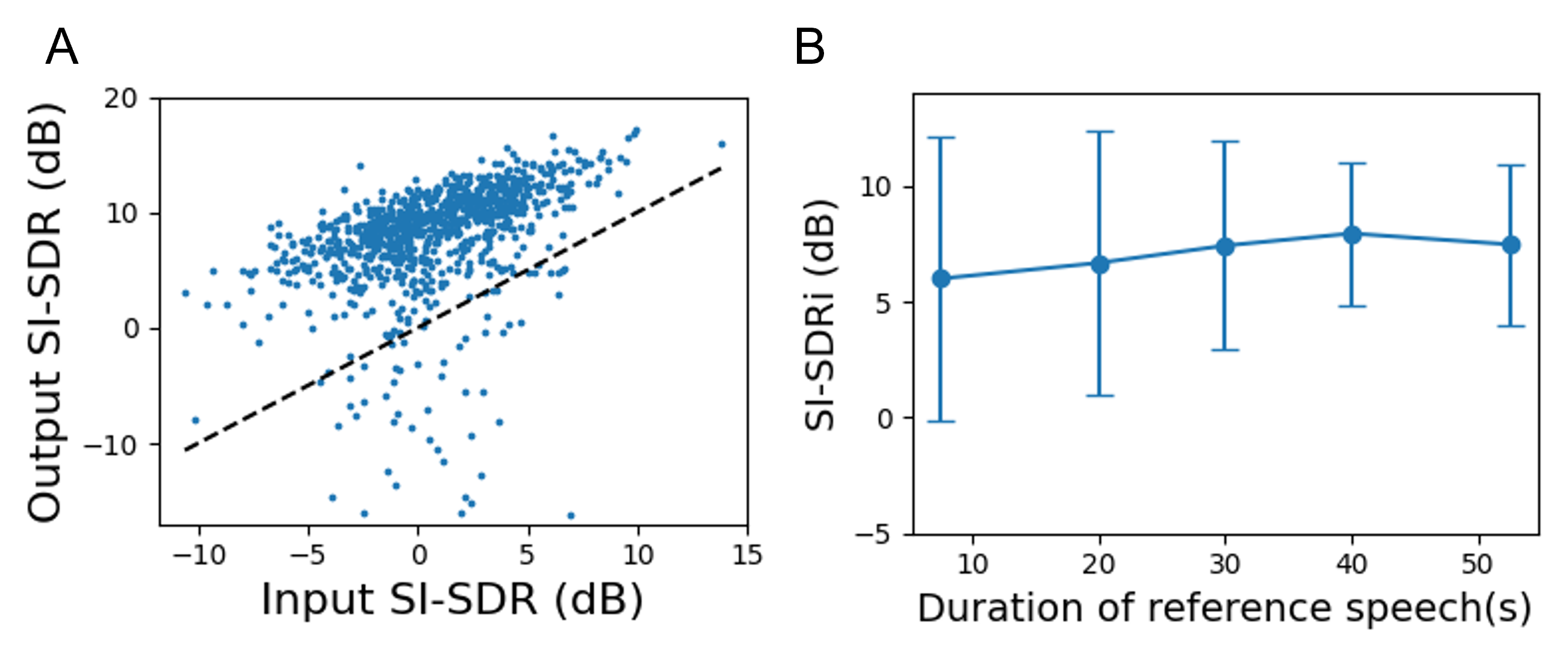}
  \vskip -0.1in
  \caption{A. Input vs output SI-SDR.  B.  Impact of the duration of the reference speaker in the conversation on SI-SDRi.}
  \label{fig:eval}
  \vskip -0.2in
\end{figure}

\section{Experiments and Results}

{\bf Datasets.} For English conversations, we used the open-source AMI corpus~\cite{kraaij2005ami}, which has timestamps, speaker ID labels, and clean audio tracks per speaker. For Mandarin conversations, we used the open-source RAMC dataset~\cite{yang2022open}.  AMI  has 2-4 speakers per  conversation with an overlap ratio of {18.6\% $\pm$ 12\%}, calculated as the ratio of overlapped speech duration to total speech duration in the conversation. The RAMC dataset has 2  speakers per conversation with an overlap ratio of 1.9\% $\pm$ 2\%. 

We randomly selected 1-minute conversation segments at a sampling rate of 16~kHz that meet two requirements:  (1) the total speech duration  in the 1-min segment exceeds 60\%, and (2) at least 2 speakers are active in each segment.  We randomly selected   one of the speakers as the reference speaker and selected their speech from {\it other}  segments as a clean enrollment example.  We used the open-source project Resemblyzer~\cite{Resemble-Ai} to compute 256-dimensional d-vectors from these clean examples, and use these as the speaker embeddings $\epsilon_0$. We   randomly sampled another 1-minute  segment from the same dataset without overlapping speakers to serve as interference.  We created three pairs of conversational datasets with mixtures. 
\vskip 0.02in\noindent$\mathcal{R}_{en}$ and $\mathcal{R}_{zh}$: These are real English and Mandarin conversation mixtures generated from AMI and RAMC. Each consists of 8000 training samples (133.3 hours), 1000 validation samples (16.6 hours), and 1000 testing samples (16.6 hours).  $\mathcal{R}_{en}$ has 2-4 speakers in the  interference signal, whereas $\mathcal{R}_{zh}$ has 2 speakers in the interference signal. Speakers in the training and validation sets do not overlap with those in the test set.

\vskip 0.02in\noindent$\mathcal{A}_{en}$ and $\mathcal{A}_{zh}$: These datasets consist of the augmented conversation mixtures and ground truth signals created using  LibriTTS~\cite{zen2019libritts} and AISHELL-3~\cite{shi2020aishell} as the open-source English and Mandarin non-conversational datasets. We set $p=50\%$ to create these datasets, which were only used for training and each have 8000 training samples (133.3 hours).

\vskip0.02in\noindent $\mathcal{S}_{stats}$ and $\mathcal{S}_{cross}$: A synthetic conversation dataset with two speakers was created using the statistical distribution for gaps and overlaps described in prior psycho-linguistics work (Fig. 2 in~\cite{frontiers}). We used speakers from LibriTTS for the audio for each speaker. We then generated the conversation mixtures  to create  $\mathcal{S}_{stats}$. We also created a cross-lingual synthetic dataset, $\mathcal{S}_{cross}$, by replacing all the Mandarin audio in the RAMC dataset  with random English speakers from the LibriTTS  dataset. The  mixtures in these datasets were used only for training and have 8000 training samples (133.3 hours) each.

 \begin{table}[t!]
  \caption{Results on mixtures of real English conversations.}
  \vskip -0.1in
  \label{tab:english}
  \centering
  \begin{tabular}{ c c c  }
    \toprule
    \multicolumn{1}{c}{\textbf{Pre-training/fine-tuning}} & 
    \multicolumn{1}{c}{\textbf{SI-SDRi}} &
    \multicolumn{1}{c}{\textbf{SNRi}} \\
    
    \midrule
     $\mathcal{R}_{en}$  & 4.74 &  7.43  \\
    $\mathcal{R}_{en}$, $\mathcal{A}_{en}$  & 4.17 & 6.78   \\
     $\mathcal{S}_{cross}, \mathcal{S}_{stats}$/$\mathcal{R}_{en}$ & 4.95  & 6.89 \\
   $\mathcal{S}_{cross}, \mathcal{S}_{stats}$/$\mathcal{R}_{en}$, $\mathcal{A}_{en}$   & \textbf{6.32}   & \textbf{7.92} \\
    \bottomrule
  \end{tabular}
    \vskip -0.2in
\end{table}

\vskip 0.04in\noindent\textbf{Training setup.} We first pre-trained a model with  $\mathcal{S}_{stats}$ and $\mathcal{S}_{cross}$ for 80 epochs. Then we fine-tuned  models with $\mathcal{A}_{en}$ and $\mathcal{R}_{en}$ for English conversations and $\mathcal{A}_{zh}$ and $\mathcal{R}_{zh}$ for Mandarin conversations. {We fine-tuned  for 20 epochs}. We used the Adam optimizer for both pre-training and fine-tuning, with gradient norm clipped to 1 and a batch size of 8.  The initial learning rate started at 0.002, halving if the validation loss did not improve in 8 epochs. We used  negative SNR as the loss function, and SI-SDRi and SNRi as evaluation metrics.  Our model configuration was as follows: STFT window size was 12.5 ms and hop size was 4 ms. The TF embedding dimension $D$ was 16, and B was 3. The pooling window $W$ was  1.25~s, equivalent to 100 TF frames, and the stride size,  $S=1.25$ s. The LSTM hidden size $H$ was 64, and the attention head number $L$ was 4.


\vskip 0.04in\noindent{\bf Results.} We  evaluated our model on the real Mandarin  conversation mixtures from $\mathcal{R}_{zh}$ which have  two speakers in each of the  target and interfering conversations.  We compared  four training configurations: (1) training on $\mathcal{R}_{zh}$ for 100 epochs, (2) training on both $\mathcal{R}_{zh}$ and $\mathcal{A}_{zh}$ for 100 epochs, (3) pre-training on $\mathcal{S}_{stats}$ and $\mathcal{S}_{cross}$ for 80 epochs and fine-tuning on $\mathcal{R}_{zh}$ for 20 epochs, (4) pre-training on $\mathcal{S}_{stats}$ and $\mathcal{S}_{cross}$ for 80 epochs and fine-tuning on $\mathcal{R}_{zh}$ and $\mathcal{A}_{zh}$  for 20 epochs  Table.~\ref{tab:mandarain} compares SNR and SI-SDR improvements  on the test set of $\mathcal{R}_{zh}$.  The average input SI-SDR was 0.67~dB and input SNR was 0.66~dB. With our data augmentation and training techniques, we achieved an SI-SDRi of 7.18 dB and SNRi of 8.19 dB, showing a 2-3 dB improvement compared to    training without any  augmentation. A paired t-test was conducted for each metric, showing a significant difference with $p<0.05$. Fig.~\ref{fig:eval}a  plots the output SI-SDR of all test samples as a function of their input SI-SDR, which shows that 93.6\% of the samples were above the zero-improvement line.  Further,  Fig.~\ref{fig:eval}b shows that  when the speech duration of the reference speaker was too short ($< 10$s) or too long ($> 50$s) within the 1-minute segments, the SI-SDR improvement drops by around 1 dB. {We also evaluated this task  with background noise. Specifically, we randomly add background noise from the WHAM! dataset  to both $\mathcal{R}_{zh}$    and  $\mathcal{A}_{zh}$ such that there was no overlap between train, test and validation splits. {We fine-tuned the model pretrained on $\mathcal{S}_{stats}$ and $\mathcal{S}_{cross}$  without noise on these new noisy mixture  datasets for 20 epochs. }We tested  it on  real Mandarin conversation mixtures with noise. We achieved a 6.15 dB SI-SDRi  and 7.68 dB SNRi (input SNR and SI-SDR were both -0.79 dB), with $p<0.05$. }

We also evaluated our model on real  English  conversation mixtures, $\mathcal{R}_{en}$, which contain 2-4 speakers in each of the target and interfering conversations. We compared four  configurations:  (1) training on $\mathcal{R}_{en}$ for 100 epochs, (2) training on $\mathcal{R}_{en}$ and $\mathcal{A}_{en}$ for 100 epochs, (3) pre-training on $\mathcal{S}_{stats}$ and $\mathcal{S}_{cross}$ for 80 epochs and fine-tuning on $\mathcal{R}_{en}$ for 20 epochs, and  (4) pre-training on $\mathcal{S}_{stats}$ and $\mathcal{S}_{cross}$ for 80 epochs and fine-tuning on $\mathcal{R}_{en}$ and $\mathcal{A}_{en}$  for 20 epochs. Table.~\ref{tab:english} compares the SNRi and SI-SDRi on the test set  of $\mathcal{R}_{en}$. The  input SI-SDR was  0.12~dB  and  input SNR 0.123~dB. Our model achieved an SI-SDRi of  6.32 dB and SNRi of 7.92 dB ($p<0.05$). 

\begin{figure}[t!]
  \centering
  \includegraphics[width=\linewidth]{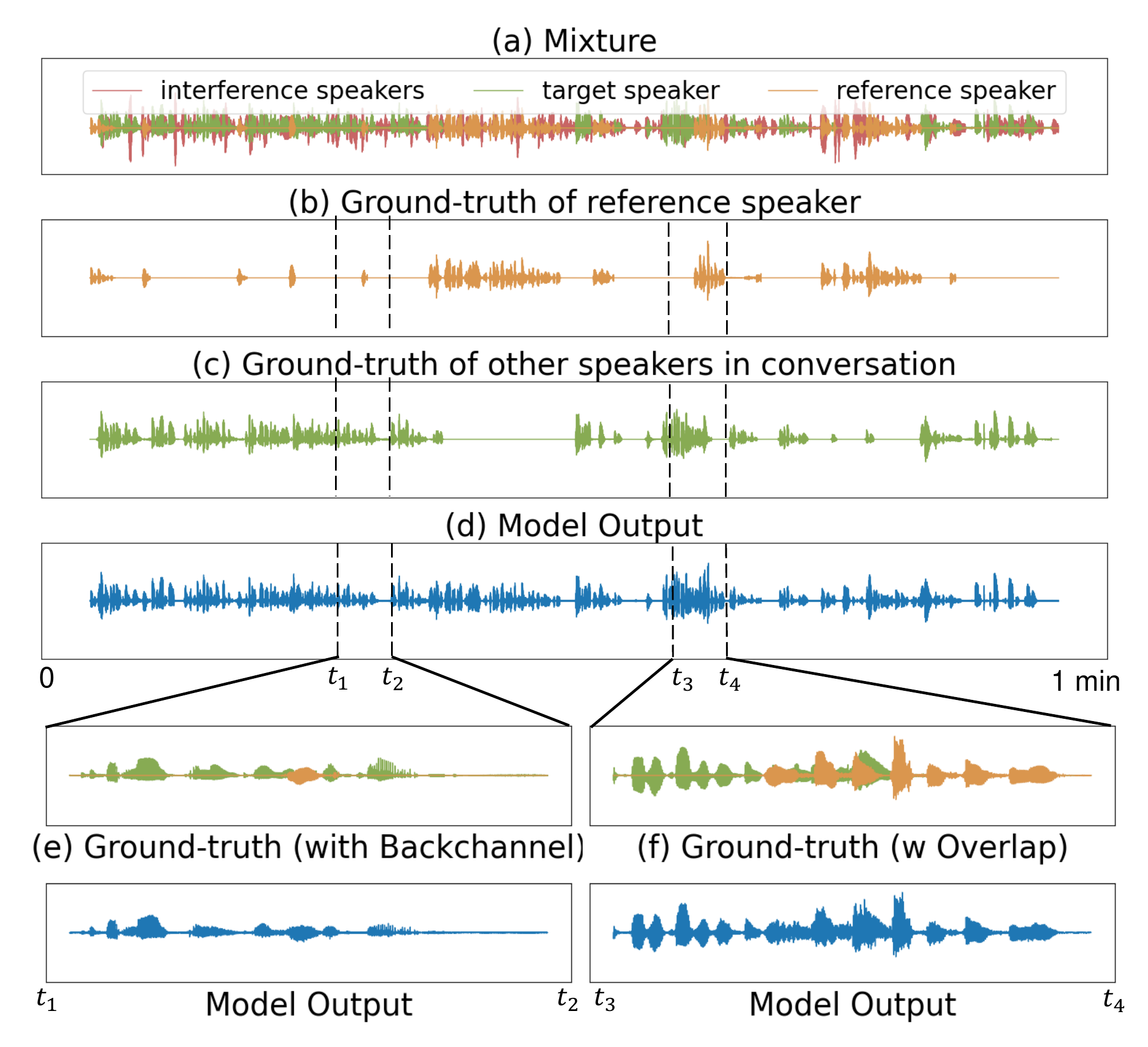}
  \vskip -0.2in
  \caption{Visualization of time-domain waveforms. (a) is the input mixture of real conversations. (b), (c) are ground-truths for speakers in the conversation. (d) is the model output. (e), (f) are the audio segments  that preserve  overlaps and back-channels.}
  \vskip -0.2in
  \label{fig:qual}
\end{figure}

Fig.~\ref{fig:qual} presents a qualitative example of real conversation waveforms. The figure shows that the model can extract the target conversation, while preserving overlaps and back-channels.

\vskip 0.04in\noindent{\bf Timing perturbation.}  To better understand our model, we perturbed the turn-taking timing patterns in the test set of  $\mathcal{R}_{zh}$.  We used  the best frozen Mandarin  model from above. We generated test samples with a reference speaker $s^0$, a target speaker $s^{conv}$, from the same conversation, and an interfering speaker $s^{inter}$, randomly chosen from another conversation, with no overlapping speakers.  We explored two perturbations: 
\begin{itemize}
\item{\it Random shifts upto $\tau$ seconds.}  Randomly time-shift utterance level speech segments independently for   reference and target speakers. Use a uniform distribution across $[-\tau, \tau]$, where negative (positive) values result in left (right)  shift. 

\item {\it Shift all speech left.}  Time-shift all speech segments for both the reference and target speakers in the conversations to the left. This would artificially  increase the overlap between the speakers and break the conversational  turn-taking dynamics.
\end{itemize}
To compute the SNRi metrics, these shifts were applied to both the mixture and the ground truth in the test set. Further, we considered a  model's  output sample to be an  incorrect target when the SNRi of the wrong conversation, $s_0 + s^{inter}$ is higher than the SNRi of the target conversation, $s_0 + s^{conv}$. The incorrect target ratio was calculated by dividing the number of incorrect output samples by the total number of test samples. 

Table.~\ref{tab:pertub} shows that randomly shifting by up to 1s resulted in only a 0.44~dB   drop in SNRi ($p<0.05$).  This is likely because  this operation does not considerably disrupt the turn-taking dynamics. However, as the random shift interval increases,  SNRi slowly drops  and the incorrect target ratio increases. {Shifting all speech left}  breaks human turn-taking dynamics,  significantly decreasing the  SNRi and increasing the incorrect target ratio. 


\begin{table}[t!]
  \caption{Timing perturbations on the test dataset. }
  \vskip -0.1in
  \label{tab:pertub}
  \centering
  \begin{tabular}{ c c c c }
    \toprule
    \multicolumn{1}{c}{\textbf{Timing}}& 
    \multicolumn{1}{c}{\textbf{Input SNR}} &
    \multicolumn{1}{c}{\textbf{SNRi}} &
    \multicolumn{1}{c}{\textbf{Incorrect  }} \\
     \textbf{perturbation}& (dB) & (dB) & \textbf{target ratio}\\
    
    \midrule
     None &  4.5 & 7.02 & 9.1\% \\
     \midrule
     Upto 1s  shifts & 4.4 & 6.58 & 10.1\% \\
     \midrule
     Upto 3s  shifts&  4.0 & 5.5 & 16.4\% \\
     \midrule
     Upto 5s  shifts &  3.8 & 4.4 & 21.7\% \\
     \midrule
     {Shift all speech left} &  4.5 & -0.12 & 53.4\% \\
    \bottomrule
  \end{tabular}
   \vskip -0.2in
\end{table}

 \setlength{\tabcolsep}{5pt}
\begin{table}[h]
\vskip -0.1in
  \caption{Ablation study and architecture  comparisons.}
  \vskip -0.1in
  \label{tab:ablation}
  \centering
  \begin{tabular}{ c c c c c c}
    \toprule
    \multicolumn{1}{c}{\textbf{Model Config}}& 
    \multicolumn{1}{c}{\textbf{SI-SDRi}} &
    \multicolumn{1}{c}{\textbf{Params  }} &
    \multicolumn{1}{c}{\textbf{RTF }} &
    \multicolumn{1}{c}{\textbf{RTF}}\\
    & & &  (CPU) &  (GPU) \\
    \midrule
     TF-GridNet &  6.46 & 2.55M &  5.87e-2 & 2.81e-3\\
     \midrule
     Mean Pool & 5.81 & 2.54M & 3.27e-2 & 1.14e-3 \\
     Max Pool &  4.72 &  2.54M & 3.14e-2 & 1.14e-3\\
     Full LSTM &  4.15 & 2.54M &  4.40e-2 & 1.02e-3\\
     Local Attention  & 5.20 & 2.39M &  2.56e-2 & 9.12e-4\\
     W = 50 &  5.60  & 2.54M &   3.41e-2 & 1.22e-3 \\
     W = 200 &  5.70  & 2.54M &  3.31e-2 & 1.12e-3 \\
    \bottomrule
  \end{tabular}
    \vskip -0.15in
\end{table}


\noindent{\bf Ablation study.} We compared the separation performance, run-time complexity, and  different model hyperparameters and architectures. For fair comparisons, we trained all models  on $\mathcal{R}_{zh}$ and $\mathcal{A}_{zh}$ for 100 epochs and tested on  $\mathcal{R}_{zh}$. We measured  inference time on both an A40  GPU  and an Intel Xeon(R) Gold 6230R CPU. We computed the real-time factor (RTF) by measuring the runtime to process a 1-min audio and divide it by a minute. We implemented the original TF-GridNet with the same hyperparameter setup as our model. We also implemented two  variants of our model: 1) Full LSTM where we removed  sparse attention and applied LSTM on the entire 1-min sequence, 2)  Local Attention, where   we replaced the temporal LSTM with a local attention module. Our model  reduced the RTF   over TF-GridNet by a factor of 1.8-2.5x, with only a 0.65~dB loss in SI-SDRi ($p<0.05$). Further, the training time reduced from 76~min per epoch for TF-GridNet to 42~min per epoch on two A40s.
Finally, we compared our network with mean and max pooling operations as well as different pooling window sizes (W), where W=100 is our default setting.

\section{Conclusion}
We introduced the novel task of target conversation extraction and demonstrated its feasibility on English and Mandarin conversation datasets. Our work has multiple limitations that present opportunities for future research.  This includes   achieving  streaming target conservation extraction, dynamic tracking of speakers entering and leaving the conversation, and incorporating  speech content and much larger language models.


\section{Acknowledgments}
The UW researchers are funded by the Moore
Inventor Fellow award \#10617, UW CoMotion fund,  and the NSF.

\bibliographystyle{IEEEtran}
\bibliography{paper}

\begin{thebibliography}{10}
\providecommand{\url}[1]{#1}
\csname url@samestyle\endcsname
\providecommand{\newblock}{\relax}
\providecommand{\bibinfo}[2]{#2}
\providecommand{\BIBentrySTDinterwordspacing}{\spaceskip=0pt\relax}
\providecommand{\BIBentryALTinterwordstretchfactor}{4}
\providecommand{\BIBentryALTinterwordspacing}{\spaceskip=\fontdimen2\font plus
\BIBentryALTinterwordstretchfactor\fontdimen3\font minus \fontdimen4\font\relax}
\providecommand{\BIBforeignlanguage}[2]{{%
\expandafter\ifx\csname l@#1\endcsname\relax
\typeout{** WARNING: IEEEtran.bst: No hyphenation pattern has been}%
\typeout{** loaded for the language `#1'. Using the pattern for}%
\typeout{** the default language instead.}%
\else
\language=\csname l@#1\endcsname
\fi
#2}}
\providecommand{\BIBdecl}{\relax}
\BIBdecl

\bibitem{frontiers}
S.~C. Levinson and F.~Torreira, ``Timing in turn-taking and its implications for processing models of language,'' \emph{Frontiers in psychology}, vol.~6, p. 731, 2015.

\bibitem{turntake1}
H.~Sacks and E.~A. Schegloff, ``(1974). a simplest systematics for the organization of turn-taking for conversation,'' \emph{Language}, vol.~50, no.~4, pp. 696--735.

\bibitem{zmolikova}
K.~Zmolikova, M.~Delcroix, T.~Ochiai, K.~Kinoshita, J.~Černocký, and D.~Yu, ``Neural target speech extraction: An overview,'' \emph{IEEE Signal Processing Magazine}, 2023.

\bibitem{turntake2}
T.~Stivers, N.~J. Enfield, P.~Brown, C.~Englert, M.~Hayashi, T.~Heinemann, G.~Hoymann, F.~Rossano, J.~P. De~Ruiter, K.-E. Yoon \emph{et~al.}, ``Universals and cultural variation in turn-taking in conversation,'' \emph{Proceedings of the National Academy of Sciences}, vol. 106, no.~26, pp. 10\,587--10\,592, 2009.

\bibitem{turntake3}
M.~Heldner and J.~Edlund, ``Pauses, gaps and overlaps in conversations,'' \emph{Journal of Phonetics}, vol.~38, no.~4, pp. 555--568, 2010.

\bibitem{yang2022open}
Z.~Yang, Y.~Chen, L.~Luo, R.~Yang, L.~Ye, G.~Cheng, J.~Xu, Y.~Jin, Q.~Zhang, P.~Zhang \emph{et~al.}, ``Open source magicdata-ramc: A rich annotated mandarin conversational (ramc) speech dataset,'' \emph{arXiv preprint arXiv:2203.16844}, 2022.

\bibitem{kraaij2005ami}
W.~Kraaij, T.~Hain, M.~Lincoln, and W.~Post, ``The ami meeting corpus,'' in \emph{Proc. International Conference on Methods and Techniques in Behavioral Research}, 2005.

\bibitem{i_vector}
O.~Glembek, L.~Burget, P.~Mat{\v{e}}jka, M.~Karafi{\'a}t, and P.~Kenny, ``Simplification and optimization of i-vector extraction,'' in \emph{ICASSP}, 2011.

\bibitem{d_vector}
Q.~Wang, C.~Downey, L.~Wan, P.~A. Mansfield, and I.~L. Moreno, ``Speaker diarization with lstm,'' in \emph{ICASSP}.\hskip 1em plus 0.5em minus 0.4em\relax IEEE, 2018.

\bibitem{speakerbeam}
K.~{\v{Z}}mol{\'\i}kov{\'a}, M.~Delcroix, K.~Kinoshita, T.~Ochiai, T.~Nakatani, L.~Burget, and J.~{\v{C}}ernock{\`y}, ``Speakerbeam: Speaker aware neural network for target speaker extraction in speech mixtures,'' \emph{IEEE Journal of Selected Topics in Signal Processing}, vol.~13, no.~4, pp. 800--814, 2019.

\bibitem{waveformer}
B.~Veluri, J.~Chan, M.~Itani, T.~Chen, T.~Yoshioka, and S.~Gollakota, ``Real-time target sound extraction,'' in \emph{ICASSP 2023}.

\bibitem{10.1145/3613904.3642057}
B.~Veluri, M.~Itani, T.~Chen, T.~Yoshioka, and S.~Gollakota, ``Look once to hear: Target speech hearing with noisy examples,'' in \emph{Proceedings of the CHI Conference on Human Factors in Computing Systems}, 2024.

\bibitem{hybridbeam}
A.~Wang, M.~Kim, H.~Zhang, and S.~Gollakota, ``Hybrid neural networks for on-device directional hearing,'' \emph{AAAI}, vol.~36, no.~10, pp. 11\,421--11\,430, 2022.

\bibitem{gu2019neural}
R.~Gu, L.~Chen, S.-X. Zhang, J.~Zheng, Y.~Xu, M.~Yu, D.~Su, Y.~Zou, and D.~Yu, ``Neural spatial filter: Target speaker speech separation assisted with directional information.'' in \emph{Interspeech}, 2019, pp. 4290--4294.

\bibitem{acousticswarm}
M.~Itani, T.~Chen, T.~Yoshioka, and S.~Gollakota, ``Creating speech zones with self-distributing acoustic swarms,'' \emph{Nature Communications}, vol.~14, 09 2023.

\bibitem{avsepformer}
J.~Lin, X.~Cai, H.~Dinkel, J.~Chen, Z.~Yan, Y.~Wang, J.~Zhang, Z.~Wu, Y.~Wang, and H.~Meng, ``Av-sepformer: Cross-attention sepformer for audio-visual target speaker extraction,'' in \emph{ICASSP 2023}.

\bibitem{liu2023separate}
X.~Liu, Q.~Kong, Y.~Zhao, H.~Liu, Y.~Yuan, Y.~Liu, R.~Xia, Y.~Wang, M.~D. Plumbley, and W.~Wang, ``Separate anything you describe,'' \emph{arXiv preprint arXiv:2308.05037}, 2023.

\bibitem{conceptbeam}
Y.~Ohishi, M.~Delcroix, T.~Ochiai, S.~Araki, D.~Takeuchi, D.~Niizumi, A.~Kimura, N.~Harada, and K.~Kashino, ``Conceptbeam: Concept driven target speech extraction,'' ser. ACM MM, 2022.

\bibitem{levinson2015timing}
S.~C. Levinson and F.~Torreira, ``Timing in turn-taking and its implications for processing models of language,'' \emph{Frontiers in psychology}, vol.~6, p. 731, 2015.

\bibitem{turn_model}
E.~Ekstedt and G.~Skantze, ``Voice activity projection: Self-supervised learning of turn-taking events,'' \emph{arXiv preprint arXiv:2205.09812}, 2022.

\bibitem{turn_eval}
E.~Ekstedt, S.~Wang, {\'E}.~Sz{\'e}kely, J.~Gustafson, and G.~Skantze, ``Automatic evaluation of turn-taking cues in conversational speech synthesis,'' \emph{arXiv preprint arXiv:2305.17971}, 2023.

\bibitem{turn_predict}
K.~Inoue, B.~Jiang, E.~Ekstedt, T.~Kawahara, and G.~Skantze, ``Real-time and continuous turn-taking prediction using voice activity projection,'' \emph{arXiv}, 2024.

\bibitem{diarization4conversation}
H.~H. Mao, S.~Li, J.~McAuley, and G.~Cottrell, ``Speech recognition and multi-speaker diarization of long conversations,'' 2020.

\bibitem{asr4conversation}
K.~Wei, Y.~Zhang, S.~Sun, L.~Xie, and L.~Ma, ``Conversational speech recognition by learning conversation-level characteristics,'' in \emph{ICASSP 2022}, 2022, pp. 6752--6756.

\bibitem{qa4conversation}
C.~You, N.~Chen, F.~Liu, S.~Ge, X.~Wu, and Y.~Zou, ``End-to-end spoken conversational question answering: Task, dataset and model,'' 2022.

\bibitem{sa4conversation}
A.~Shenoy and A.~Sardana, ``Multilogue-net: A context-aware rnn for multi-modal emotion detection and sentiment analysis in conversation,'' in \emph{Second Grand-Challenge and Workshop on Multimodal Language (Challenge-HML)}, 2020.

\bibitem{sdu4conversation}
T.~Yu, H.~Gao, T.-E. Lin, M.~Yang, Y.~Wu, W.~Ma, C.~Wang, F.~Huang, and Y.~Li, ``Speech-text dialog pre-training for spoken dialog understanding with explicit cross-modal alignment,'' 2023.

\bibitem{wang2023tf}
Z.-Q. Wang, S.~Cornell, S.~Choi, Y.~Lee, B.-Y. Kim, and S.~Watanabe, ``Tf-gridnet: Integrating full-and sub-band modeling for speech separation,'' \emph{IEEE/ACM Transactions on Audio, Speech, and Language Processing}, 2023.

\bibitem{vaswani2017attention}
A.~Vaswani, N.~Shazeer, N.~Parmar, J.~Uszkoreit, L.~Jones, A.~N. Gomez, {\L}.~Kaiser, and I.~Polosukhin, ``Attention is all you need,'' \emph{Advances in neural information processing systems}, vol.~30, 2017.

\bibitem{longformer}
I.~Beltagy, M.~E. Peters, and A.~Cohan, ``Longformer: The long-document transformer,'' \emph{arXiv}, 2020.

\bibitem{poolingformer}
H.~Zhang, Y.~Gong, Y.~Shen, W.~Li, J.~Lv, N.~Duan, and W.~Chen, ``Poolingformer: Long document modeling with pooling attention,'' in \emph{International Conference on Machine Learning}.\hskip 1em plus 0.5em minus 0.4em\relax PMLR, 2021, pp. 12\,437--12\,446.

\bibitem{resepformer}
C.~Subakan, M.~Ravanelli, S.~Cornell, F.~Lepoutre, and F.~Grondin, ``Resource-efficient separation transformer,'' \emph{arXiv preprint arXiv:2206.09507}, 2022.

\bibitem{speechLongformer}
B.~Alastruey, G.~I. G{\'a}llego, and M.~R. Costa-juss{\`a}, ``Efficient transformer for direct speech translation,'' \emph{arXiv}, 2021.

\bibitem{perez2018film}
E.~Perez, F.~Strub, H.~De~Vries, V.~Dumoulin, and A.~Courville, ``Film: Visual reasoning with a general conditioning layer,'' in \emph{AAAI}, vol.~32, no.~1, 2018.

\bibitem{landini2022simulated}
F.~Landini, A.~Lozano-Diez, M.~Diez, and L.~Burget, ``From simulated mixtures to simulated conversations as training data for end-to-end neural diarization,'' \emph{arXiv}, 2022.

\bibitem{yang2023simulating}
M.~Yang, N.~Kanda, X.~Wang, J.~Wu, S.~Sivasankaran, Z.~Chen, J.~Li, and T.~Yoshioka, ``Simulating realistic speech overlaps improves multi-talker asr,'' in \emph{ICASSP 2023}.

\bibitem{kanda2022streaming}
N.~Kanda, J.~Wu, Y.~Wu, X.~Xiao, Z.~Meng, X.~Wang, Y.~Gaur, Z.~Chen, J.~Li, and T.~Yoshioka, ``Streaming multi-talker asr with token-level serialized output training,'' \emph{arXiv}, 2022.

\bibitem{Resemble-Ai}
\BIBentryALTinterwordspacing
Resemble-Ai, ``Resemble-ai/resemblyzer: A python package to analyze and compare voices with deep learning,'' 2019. [Online]. Available: \url{https://github.com/resemble-ai/Resemblyzer}
\BIBentrySTDinterwordspacing

\bibitem{zen2019libritts}
H.~Zen, V.~Dang, R.~Clark, Y.~Zhang, R.~J. Weiss, Y.~Jia, Z.~Chen, and Y.~Wu, ``Libritts: A corpus derived from librispeech for text-to-speech,'' \emph{arXiv preprint arXiv:1904.02882}, 2019.

\bibitem{shi2020aishell}
Y.~Shi, H.~Bu, X.~Xu, S.~Zhang, and M.~Li, ``Aishell-3: A multi-speaker mandarin tts corpus and the baselines,'' \emph{arXiv preprint arXiv:2010.11567}, 2020.

\end{thebibliography}

\end{document}